\documentclass[11pt]{article}

\usepackage[preprint]{acl}
\usepackage{times}
\usepackage{latexsym}

\usepackage[T1]{fontenc}

\usepackage[utf8]{inputenc}

\usepackage{microtype}

\usepackage{inconsolata}

\usepackage{graphicx}

\usepackage{booktabs}

\title{Beyond Aggregate WER: In-domain Adaptation, Learner Speech, Code-Switching, and Speaker Attribution in Online Spanish Learning Classrooms}

\author{
  Olga Kellert$^{1}$\thanks{Equal contribution as first authors.}\thanks{Corresponding author.}\quad
  Samhitha Harish$^{1}$\footnotemark[1]\quad
  Johnatan E Bonilla$^{2}$\footnotemark[1]\quad
  Nemika Tyagi$^{1}$\thanks{Equal contribution as second authors.}\quad\\
  \textbf{Phanos Kareen}$^{1}$\footnotemark[3]\quad
  \textbf{Nelvin Licona-Guevara}$^{1}$\footnotemark[3] \\\\
  $^1$Arizona State University \quad $^2$Humboldt-Universit\"at zu Berlin \\
  \small{\texttt{\{olga.kellert, sharish5, ntyagi8, kphanos, nliconag\}@asu.edu}} \\
  \small{\texttt{j.bonilla@hu-berlin.de}}
}

\begin{document}
\maketitle

\begin{abstract}
    Automatic speech recognition (ASR) increasingly supplies transcripts for educational NLP, but aggregate word error rate (WER) can conceal errors concentrated in learner speech, multilingual turns, and speaker attribution. We compare platform-integrated, open-weight, adapted, and commercial ASR configurations on held-out sessions of online L2 Spanish  interaction between L1 English learners and their instructors and tutors. Because English is the learner's shared first language, the interactions contain Spanish-only, English-only, and code-switched turns. Evaluation combines overall WER and character error rate (CER), WER disaggregated by speaker role and language context, and diarization error rate (DER). We also adapt Whisper large-v3 and MOSS-Transcribe-Diarize with LoRA using human-corrected in-domain classroom data. MOSS adaptation produced consistent improvements across speaker roles and language contexts, whereas Whisper showed smaller gains. Domain match proved more important than data volume: Larger out-of-domain bilingual and learner corpora did not outperform the smaller classroom corpus, where learner speech and code-switching are embedded in instructor--learner interaction. The open-weight MOSS adapter was competitive with commercial Muse Voice Transcribe on WER and CER and achieved lower system-provided DER on the recordings eligible for each system. Learner, English-only, and code-switched speech remained more difficult than instructor and Spanish-only speech. The findings demonstrate the value of condition-aware evaluation and interactionally matched data for bilingual educational ASR.

\end{abstract}

\section{Introduction}
\label{sec:introduction}

Automatic speech recognition (ASR) increasingly supplies transcripts for educational technologies and downstream NLP applications, including accessibility tools, spoken learner-error detection, proficiency assessment, feedback generation, and classroom-discourse analysis, so recognition errors affect not only readability but also any measurement or prediction based on the transcript.

Authentic second-language (L2) classrooms are a demanding recognition environment. Learner speech may contain developing pronunciation, non-target forms, hesitations, repetitions, and repairs, and recordings involve multiple speakers, variable microphones, rapid turn-taking, and switches between the target language and another shared language. Instructors and learners also differ in proficiency, speaking time, and interactional role, so aggregate word error rate (WER) can obscure poorer performance on learner, English-only, or code-switched speech. Moreover, a lexically accurate transcript may be unsuitable for participation analysis if speech is assigned to the wrong speaker.

Prior research has documented substantial ASR variation across educational recordings, accents, speakers, and recording conditions \citep{southwell2022challenges,kuhn2023measuring,graham2024evaluating}. Matched learner data can improve recognition \citep{kim2024learnervoice}, but conventional ASR may normalize precisely the learner forms needed for linguistic analysis or feedback \citep{michot2024errorpreserving,koizumi2025accuracy}. Code-switching introduces further language-confusion errors even in multilingual systems \citep{hillah2024tevienes,yang2025spt}, and speaker attribution is a related but separate evaluation problem \citep{vonneumann2023meeteval}. These dimensions have generally been studied separately rather than within a single evaluation of authentic bilingual instructional interaction.

\begin{figure*}[t]
    \centering
    \includegraphics[width=\textwidth]{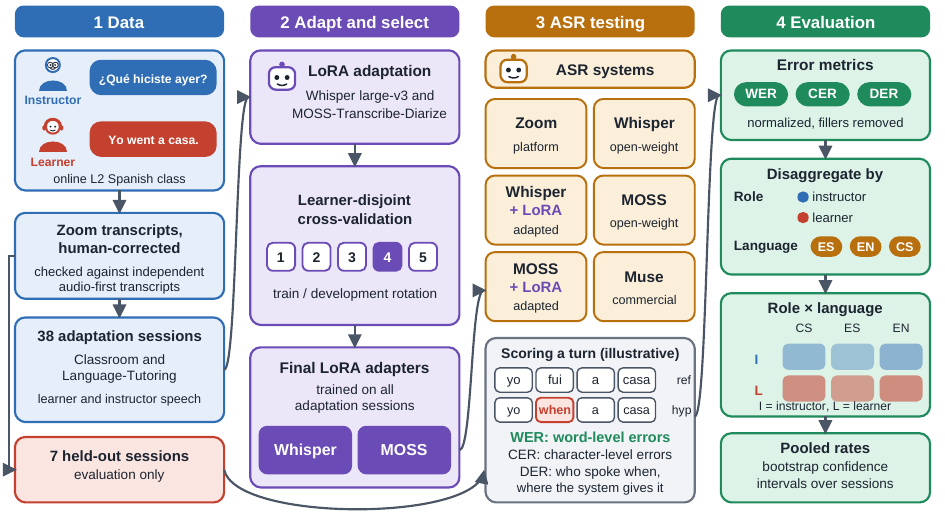}
    \caption{Study pipeline. Whisper and MOSS are adapted with LoRA on 38 human-corrected Zoom sessions, with the recipe fixed by five-fold cross-validation. Six systems are scored on seven held-out sessions against human-corrected references.}
    \label{fig:asr_pipeline}
\end{figure*}

We evaluate six ASR configurations on synchronous online Spanish L2 interaction: Zoom automated captions, Whisper large-v3 \citep{radford2023robust}, a Whisper large-v3 adapter trained with in-domain low-rank adaptation \citep[LoRA;][]{hu2022lora}, MOSS-Transcribe-Diarize \citep{yu2026moss}, MOSS with LoRA adaptation, and Muse Voice Transcribe \citep{meta2026muse}. The learners were L1 English speakers, and English was used alongside Spanish in learner production, explanations, and classroom management, so the interaction includes Spanish-only, English-only, and code-switched turns. Whisper and MOSS were adapted on 38 human-corrected Zoom transcripts (15.6 hours), with the recipe and training budget fixed by five-fold, learner-disjoint cross-validation before evaluation (Appendix~\ref{app:adaptation}). Evaluation uses seven held-out sessions from graded university classes and ungraded Language-Tutoring conversations. We report pooled WER and character error rate (CER), disaggregate WER by instructor versus learner role and by language context, and evaluate speaker attribution separately using diarization error rate (DER). Figure~\ref{fig:asr_pipeline} summarizes the design. Adapters and de-identified transcripts will be released on request only; scoring code is in the supplementary material (Appendix~\ref{app:supplementary}).

The study makes three principal contributions:

\begin{itemize}
    \item We provide a condition-aware comparison of platform-integrated, open-weight, adapted, and commercial ASR configurations on authentic bilingual interaction from online L2 instructional settings.
    \item We adapt Whisper and MOSS on the same in-domain corpus: gains are substantial and consistent for MOSS across roles and language contexts, modest for Whisper, and not improved by larger out-of-domain bilingual and learner corpora.
    \item We show that Muse and the MOSS adapter form a stable leading pair across speaker roles and language contexts, while learner and multilingual speech remain difficult and speaker attribution yields a different ranking, led by the open-weight MOSS adapter.
\end{itemize}

\section{Related Work}
\label{sec:related_work}

\subsection{L2 and Classroom ASR}

ASR performance on non-native speech varies with accent, first-language background, proficiency, and speech type \citep{graham2024evaluating}, and spontaneous learner speech adds developing pronunciation, non-target forms, hesitations, repetitions, and repairs. \citet{kim2024learnervoice} showed that fine-tuning Whisper on spontaneous non-native English substantially reduced WER, demonstrating the value of matched learner data.

Lower WER, however, does not necessarily yield a more useful transcript for learner-focused NLP, because ASR systems may normalize a learner's production to its presumed target form. \citet{michot2024errorpreserving} therefore distinguish recognition accuracy from learner-error preservation, and \citet{koizumi2025accuracy} documents similar normalization in adult L2 speech. These findings motivate preserving comprehensible non-target forms in our references.

Classroom recordings add background noise, overlapping talk, and multiple speakers. \citet{southwell2022challenges} found high error rates on middle-school collaborative speech, with errors affecting downstream discourse analysis. \citet{attia2025weak} demonstrated that approximately 5,000 hours of noisy teacher--student transcripts could support classroom ASR adaptation when followed by accurate in-domain fine-tuning. However, prior studies either focus on student speech or treat teacher and student speech as one classroom domain, and do not directly compare instructor and learner WER within the same interactions. We address this gap through role-conditioned evaluation.

\subsection{Code-Switched ASR and Adaptation}

Multilingual pretraining does not eliminate language confusion in code-switched speech. Material in the locally non-dominant language may be deleted, translated, or replaced with words from the dominant language. For Spanish--English speech, \citet{hillah2024tevienes} showed that jointly fine-tuning Whisper for language identification and transcription improved code-switched recognition. \citet{dhawan2023unified} incorporated token-level language identification into English--Spanish and English--Hindi ASR, while \citet{yang2023adapting} found that relatively small quantities of matched Mandarin--English data substantially improved Whisper.

These studies show the value of language-aware adaptation, but they primarily use established bilingual corpora rather than instructional interaction involving learners, instructors, short switches, fragments, and repairs. We therefore evaluate Spanish-only, English-only, and code-switched speech separately, including their interaction with speaker role, and test whether LoRA adaptation \citep{hu2022lora} benefits these conditions consistently.

\subsection{Role-Conditioned and Speaker-Aware Evaluation}

Aggregate WER can conceal differences across speakers and conditions. \citet{imai-etal-2025-evaluating} found substantial variation across conversational datasets, acoustic conditions, and speaker characteristics, while \citet{rai2025asrfairbench} showed that performance disparities vary across demographic groups. Instructional role introduces another potentially important distinction because instructor and learner speech support different educational and NLP applications.

Lexical accuracy also does not indicate whether speech was attributed to the correct participant. \citet{vonneumann2023meeteval} provide speaker-aware measures that jointly evaluate transcription and speaker assignment, whereas DER isolates speaker confusion, missed speech, and false alarms. The two perspectives are complementary: a system may recognize the words accurately while assigning them to the wrong speaker.

Previous research has therefore examined learner speech, classroom adaptation, code-switching, conditioned evaluation, and diarization largely as separate problems; the present study combines them.

\section{Dataset}
\label{sec:dataset}

\subsection{Corpus and Data Splits}
\label{sec:data_splits}

The corpus consists of synchronous online Spanish L2 interaction recorded over two academic terms. It includes graded, task-oriented classes led by trained instructors and ungraded \textit{Language-Tutoring} conversations led by native Spanish speakers. Classroom sessions contained structured activities, assessment, corrective feedback, presentations, and media playback; Language-Tutoring sessions were organized around tutor-selected conversation topics.

Whisper and MOSS were adapted on 38 sessions representing 15.6 hours of interaction, and evaluation used seven sessions excluded from adaptation, representing approximately 3.51 hours (Table~\ref{tab:dataset_composition}). The adaptation sessions were approximately 25 minutes long. Because the corpus consisted predominantly of beginner-level, instructor-led interaction, instructor speech generally exceeded learner speech, and Spanish was the primary language.

The 38 adaptation sessions contained 101,678 reference words before filler removal, 76,998 (75.7\%) from classroom sessions and 24,680 (24.3\%) from Language-Tutoring sessions. Learners produced about 24\% of the words overall, 21.2\% in the classroom sessions and 34.3\% in Language-Tutoring. By language context, 58.3\% of the words occurred in Spanish-only turns, 33.1\% in code-switched turns, and 8.6\% in English-only turns (Section~\ref{sec:language_identification}). The share of English words per session ranged from 1.1\% to 62.7\%, with a median of 15.6\%. Instructors and learners used English at almost the same rate, 21.5\% and 21.8\% of their words. The evaluation sessions contained a smaller share of code-switched words (18.3\%, Section~\ref{sec:language_results}).

%
%

\begin{table*}[t]
\centering
\small
\begin{tabular}{lcccl}
\hline
\textbf{Split} &
\textbf{Total} &
\textbf{Classroom} &
\textbf{Language-Tutoring} &
\textbf{Course levels} \\
\hline
Adaptation & 38 & 25 & 13 &
31 beginner; 7 intermediate \\
Evaluation & 7 & 5 & 2 &
4 beginner; 2 intermediate; 1 advanced \\
\hline
\end{tabular}
\caption{Number of sessions in the adaptation and evaluation sets. Splits were defined at the session level.}
\label{tab:dataset_composition}
\end{table*}

\subsection{Reference Transcripts}
\label{sec:reference}

Transcribers proficient in English and Spanish produced the primary transcripts by manually correcting Zoom-generated transcripts. Transcribers corrected recognition errors, added omitted speech, removed spurious material, and verified speaker attribution, code-switching, repetitions, incomplete words, and overlapping speech. Unintelligible material was marked explicitly. Comprehensible learner pronunciations and non-target forms were preserved rather than normalized because such forms are relevant to learner-focused NLP. Complete annotation conventions are provided in Appendix~\ref{app:transcription_guidelines}.

To test the correction method, independent annotators transcribed a small audio subset directly from the recordings, without Zoom or other ASR output (Appendix~\ref{app:transcription_guidelines}). The human-corrected transcripts agreed closely with these audio-first transcripts, indicating that correction introduces no systematic differences from what was spoken (Appendix~\ref{app:reference_verification}). We therefore use the human-corrected Zoom transcripts, which cover all evaluation sessions.

\section{Methods}
\label{sec:methods}

\subsection{ASR Systems and Adaptation}
\label{sec:asr_systems}
We evaluated Zoom automated captions, Whisper large-v3 \citep{radford2023robust}, MOSS-Transcribe-Diarize \citep{yu2026moss}, and Muse Voice Transcribe \citep{meta2026muse}, a commercial service accessed through its API, together with Whisper large-v3 adapter and MOSS adapter, the same two models with an in-domain LoRA adapter \citep{hu2022lora} trained on the 38 sessions described in Section~\ref{sec:data_splits}. We used LoRA because the adapted models must continue to recognize English as well as Spanish, and full fine-tuning on 15.6 hours of speech can overwrite that ability: LoRA forgets less of the original model than full fine-tuning \citep{biderman2024lora}, and it has been used to add languages or domains to Whisper without degrading the languages already covered \citep{song2024lorawhisper,li2025loraexperts}. MOSS was selected as the second open-weight model because it transcribes and attributes speakers in a single pass over a complete recording.

\paragraph{Adapter configuration.}
Both adapters used rank $r = 64$, scaling parameter $\alpha = 128$, and dropout 0.1. In MOSS, LoRA was applied to every attention and feed-forward projection of the audio encoder and the language decoder, giving 68.7 million trainable parameters (7.5\% of the model), while the projector between the two components remained frozen. In Whisper, LoRA was applied to the same projections, giving approximately 115 million trainable parameters (7.4\%). Both were trained with PEFT and the Hugging Face Trainer on one NVIDIA A100 80~GB GPU, at a learning rate of $5\times10^{-5}$ selected on development data. A MOSS training example was a complete recording and a Whisper example was a window of at most 30 seconds of contiguous audio, a difference that carried over to batch size, schedule, and checkpoint selection (Appendix~\ref{app:training_config}).

\paragraph{Training budget and model selection.}
The training budget was selected by five-fold cross-validation over the 38 adaptation sessions, with sessions grouped so that no learner appeared in both the training and the test portion of a fold; each fold trained on approximately 28 sessions and selected its checkpoint on two development sessions. For MOSS, each fold trained for six epochs; four of the five folds selected the final checkpoint, 252 to 264 steps, and one selected epoch 4, 176 steps, the two differing by less than one point of development WER in every fold. The final adapter was therefore trained for 264 steps, the budget selected by most folds, which on the 57 examples of the full set is 4.6 passes. For Whisper, the budget was 330 steps, the mean of the best steps of the five folds. The final checkpoint, step 275, was chosen on windows of its own training data, without evaluation sessions (Appendix~\ref{app:training_config}). Appendix~\ref{app:cv} reports the cross-validated effect. Alternative adapter configurations, explicit language conditioning, and the addition of external bilingual \citep{deuchar2014miami} and learner \citep{mitchell2008splloc} corpora were compared on development sessions; none improved on this recipe (Appendices~\ref{app:variants} to~\ref{app:external}).

\subsection{Inference and Processing}
\label{sec:inference}

Whisper processed reference-defined speaker turns, whereas MOSS and Muse processed longer segments or complete recordings; Zoom retained its original meeting segmentation. Whisper therefore received reference-derived turn boundaries, which removes segmentation errors from its score, and the results compare complete system configurations rather than model architectures alone. Whisper detected the language automatically at inference rather than being forced to Spanish. Outputs were ordered temporally and aligned with each reference as a continuous sequence.

References and hypotheses underwent identical normalization: text was lowercased, punctuation and nonevaluated annotations were removed, and numerical expressions were converted to spoken forms. Diacritics and markers representing contractions or incomplete words were retained. For the primary analysis, standardized filler variants were removed from both references and hypotheses because filler-only turns could not be assigned reliable language-context labels and destabilized sequence alignment, while repetitions, repairs, fragments, and incomplete words were preserved. The complete filler inventory is provided in Appendix~\ref{app:filler-inventory}.

\subsection{Language Classification}
\label{sec:language_identification}

The classifier assigned each token to Spanish or English by comparing its Zipf frequency in the two languages \citep{speer2022wordfreq}. Tokens whose frequency difference fell below a margin of 0.75 were classified as ambiguous. Named entities and established loanwords were identified using curated lists; remaining tokens without sufficient evidence were classified as unknown. Ambiguous tokens did not contribute evidence for either language.

Turns containing Spanish but no English evidence were classified as Spanish-only, those containing English but no Spanish evidence as English-only, and those containing evidence from both languages as code-switched. Turns with no evidence for either language were excluded from the language-context analyses. Manual review of one complete transcript found 265 of 271 turn classifications to be correct (97.79\%). The tagger and accompanying word lists are provided in the supplementary material (Appendix~\ref{app:supplementary});



\subsection{Transcription Evaluation}
\label{sec:asr_scoring}

We report WER and CER based on standard edit-distance alignment \citep{morris04_interspeech}. Error counts and reference units were summed across recordings before each rate was calculated; session-level percentages were not averaged. Intervals containing external media without corresponding reference speech were excluded. The normalization and scoring scripts are provided in the supplementary material (Appendix~\ref{app:supplementary}).

WER and CER were calculated overall, and WER was additionally disaggregated by speaker role and language context. Substitutions and deletions inherited the category of the corresponding reference token, while insertions were assigned to the preceding reference category. Reference-word and turn counts accompany the condition-specific results.

\paragraph{Uncertainty estimation.}
We estimated uncertainty in overall WER using a session-level bootstrap, treating the seven evaluation sessions as the sampling units \citep{koehn2004significance}. For each of 10,000 bootstrap replicates, seven sessions were sampled with replacement, and pooled WER was recomputed by dividing the total number of substitutions, deletions, and insertions across the sampled sessions by the corresponding total number of reference words. We report 95\% percentile-bootstrap confidence intervals based on the 2.5th and 97.5th percentiles of the resulting WER distributions.

\subsection{Diarization Evaluation}
\label{sec:diarization_method}

DER was computed using \texttt{pyannote.metrics} \citep{bredin2017pyannote}, with a 0.25-second collar in total (0.125 seconds on each side of every reference boundary) and overlapping speech retained. System-provided speaker attribution was evaluated for  MOSS base, MOSS adapter, and Muse; Zoom's participant labels were scored separately (Section~\ref{sec:diarization_results}). We also evaluated pyannote-3.1 \citep{plaquet2023powerset,bredin2023pyannote}, pyannote-community-1, DiariZen \citep{han2025leveraging}, NeMo Sortformer \citep{park2025sortformer}, and Nemotron-3-Diarization \citep{nvidia2026nemotron3diarization}.

External diarizers processed complete recordings, and their predictions were transferred to ASR turn boundaries using majority temporal overlap. NeMo Sortformer processed six-minute chunks, with speaker identities linked across chunks using embedding similarity.

Outputs reproducing reference boundaries or speaker labels were excluded to prevent circular evaluation. Whisper was therefore ineligible because it processed reference-defined speaker turns and made no independent attribution decisions. A system was eligible on a recording only if its output on that recording did not reproduce reference boundaries or speaker labels, and DER was pooled across each system's eligible recordings by summing error durations and reference speech time.

\section{Results}
\label{sec:results}

\subsection{Transcription Performance}
\label{sec:transcription_results}


All results use filler-cleaned text and pooled rates (Section~\ref{sec:asr_scoring}). The instructor category includes classroom instructors and the tutors of the Language-Tutoring sessions. Speaker-role analyses use reference speaker labels and are therefore independent of automatic diarization.

\subsubsection{Overall Performance}

Table~\ref{tab:overall_error} reports pooled WER and CER across the seven evaluation sessions.

\begin{table}[!t]
\centering
\small
\setlength{\tabcolsep}{4pt}
\begin{tabular}{lccc}
\hline
\textbf{System} & \textbf{WER} & \textbf{95\% CI} & \textbf{CER} \\
\hline
Zoom captions   & 36.00 & {[}32.67, 40.17{]} & 31.62 \\
Whisper large-v3    & 34.50 & {[}31.97, 37.89{]} & 28.72 \\
Whisper large-v3 adapter & 32.30 & {[}29.24, 35.74{]} & 27.00 \\
MOSS base       & 26.52 & {[}21.48, 31.85{]} & 19.08 \\
MOSS adapter    & 16.39 & {[}13.46, 20.41{]} & 12.71 \\
Muse            & \textbf{15.59} & {[}13.07, 19.01{]} & \textbf{10.43} \\
\hline
\end{tabular}
\caption{Overall WER and CER across the seven evaluation sessions. WER is followed by its 95\% session-bootstrap confidence interval. All values are percentages; lower is better.}
\label{tab:overall_error}
\end{table}

At the point-estimate level, Muse achieved the lowest overall WER and CER, at 15.59\% and 10.43\%, respectively. The MOSS adapter ranked second, with 16.39\% WER and 12.71\% CER. The session-bootstrap WER intervals for Muse and the MOSS adapter overlap substantially, and we therefore treat the two as a leading pair rather than ranking one above the other.

Adaptation improved both model families, although the effect was substantially larger for MOSS. Whisper adaptation reduced WER from 34.50\% to 32.30\% (2.20 percentage points, or 6.4\% relative) and CER from 28.72\% to 27.00\% (1.72 points, or 6.0\% relative). MOSS adaptation reduced WER from 26.52\% to 16.39\% (10.13 points, or 38.2\% relative) and CER from 19.08\% to 12.71\% (6.37 points, or 33.4\% relative). The gain came from matched data rather than from volume. Under learner-disjoint cross-validation over the 38 adaptation sessions, MOSS adaptation reduced WER by 3.14 points with a paired interval that excluded zero, and seven training mixtures that added bilingual conversation or out-of-classroom learner speech, one of those corpora several times larger than our adaptation set, never improved on the classroom sessions alone (Appendices~\ref{app:cv} and~\ref{app:external}).

\subsubsection{Performance by Speaker Role}

Table~\ref{tab:role_wer} reports WER separately for instructor and learner speech. Instructors produced 1,208 turns and 12,194 reference words (72.3\% of the role-labeled words), whereas learners produced 843 turns and 4,672 words (27.7\%). The learner--instructor gap is expressed in percentage points rather than as a ratio.

\begin{table}[!t]
\centering
\small
\setlength{\tabcolsep}{3pt}
\resizebox{\columnwidth}{!}{%
\begin{tabular}{lccc}
\hline
\textbf{System} & \textbf{Instructor} & \textbf{Learner} & \textbf{Gap (pp)} \\
\hline
Zoom captions   & 29.00 & 54.27 & 25.27 \\
Whisper large-v3    & 28.20 & 50.94 & 22.74 \\
Whisper large-v3 adapter & 26.50 & 47.44 & 20.94 \\
MOSS base       & 21.80 & 38.84 & 17.04 \\
MOSS adapter    & 13.46 & 24.04 & 10.58 \\
Muse            & \textbf{12.80} & \textbf{22.87} & \textbf{10.07} \\
\hline
\end{tabular}%
}
\caption{WER by speaker role. Values are percentages; the learner--instructor gap is reported in percentage points.}
\label{tab:role_wer}
\end{table}

Muse obtained the lowest WER for both speaker roles, with 12.80\% for instructors and 22.87\% for learners. The MOSS adapter followed closely, achieving 13.46\% WER on instructor speech and 24.04\% on learner speech. Across all systems, learner speech was more challenging than instructor speech, although the magnitude of this difference varied substantially across models. Zoom exhibited the largest learner--instructor gap at 25.27 percentage points, followed by Whisper large-v3 at 22.74 points and Whisper adapter at 20.94 points.

Adaptation improved both speaker roles for both models, with a larger absolute improvement for learner speech. For Whisper, instructor WER decreased from 28.20\% to 26.50\% (1.70 percentage points) and learner WER from 50.94\% to 47.44\% (3.50 points), narrowing the learner--instructor gap from 22.74 to 20.94 points. MOSS adaptation produced substantially larger improvements: instructor WER decreased from 21.80\% to 13.46\% (8.34 points) and learner WER from 38.84\% to 24.04\% (14.80 points), narrowing the gap from 17.04 to 10.58 points.


\subsubsection{Performance by Language Context}
\label{sec:language_results}

Table~\ref{tab:language_wer} reports WER for Spanish-only, English-only, and code-switched speech. The language-context subset covers all 2,051 role-labeled turns: 1,635 Spanish-only turns with 13,064 reference words (77.5\%), 251 code-switched turns with 3,086 words (18.3\%), and 165 English-only turns with 716 words (4.2\%).

\begin{table}[!t]
\centering
\small
\setlength{\tabcolsep}{6pt}
\begin{tabular}{lccc}
\hline
\textbf{System} & \textbf{CS} & \textbf{Spanish} & \textbf{English} \\
\hline
Zoom captions   & 47.20 & 32.34 & 54.55 \\
Whisper large-v3    & 44.83 & 31.03 & 53.34 \\
Whisper large-v3 adapter & 41.67 & 29.09 & 50.58 \\
MOSS base       & 33.54 & 24.36 & 35.67 \\
MOSS adapter    & 21.44 & 14.77 & 24.14 \\
Muse            & \textbf{20.41} & \textbf{14.05} & \textbf{22.93} \\
\hline
\end{tabular}
\caption{WER by language context. Values are percentages; CS denotes code-switched speech.}
\label{tab:language_wer}
\end{table}

System rankings were generally consistent across language contexts. Muse obtained the lowest WER for code-switched and English-only speech, at 20.41\% and 22.93\%, respectively, and also obtained the lowest Spanish-only WER at 14.05\%. The MOSS adapter followed closely across all three language contexts, with 21.44\% WER for code-switched speech, 14.77\% for Spanish-only speech, and 24.14\% for English-only speech. Zoom captions showed the highest error rates across the three conditions, reaching 47.20\% on code-switched speech and 54.55\% on English-only speech.

Adaptation effects were more strongly differentiated by model configuration than by speaker role or language context. MOSS adaptation produced substantial improvements across both roles and all three language contexts whereas Whisper adaptation produced consistently smaller gains. The joint role--language analysis showed the same pattern: Muse obtained the lowest WER in all six conditions, followed closely by the MOSS adapter, and both adapters improved all six conditions, with considerably larger reductions for MOSS (Appendix~\ref{app:role_language_wer}).

Both adapters were trained on the same 38-session in-domain corpus, so the contrast cannot be attributed to differences in adaptation-data quantity or composition. The configurations nevertheless differed in architecture, training segmentation, and inference procedure: Whisper processed shorter, reference-defined segments, whereas MOSS was trained and evaluated on longer recordings or complete sessions. The results therefore establish a configuration-level difference in adaptation effectiveness; development experiments point to the decoding regime as one such factor. In cross-validation, the Whisper adapter gained 4.54 points when decoded in 30-second windows, its training regime, but lost the gain under long-form decoding, whereas MOSS, which decodes complete recordings, kept its gain (Appendix~\ref{app:conditioning}).

Because instructor speech and Spanish-only speech constitute most of the evaluation words, aggregate WER could primarily reflect performance on these relatively easier majority conditions. The disaggregated results show instead that the advantages of Muse and the MOSS adapter extend to learner, code-switched, and English-only speech. At the same time, substantial gaps remain even for these leading configurations: learner WER is approximately 10 percentage points higher than instructor WER, while code-switched and English-only WER exceed Spanish-only WER by approximately 6--9 points. The condition-specific analyses therefore establish the robustness of the leading configurations while identifying where recognition remains most difficult.


\subsection{Diarization Performance}
\label{sec:diarization_results}

Diarization was evaluated separately from transcription; consequently, the speaker labels used in the preceding role-conditioned WER analyses were always taken from the reference rather than from automatic diarization output.


Table~\ref{tab:native_der} reports pooled DER for the long-form ASR systems evaluated using their system-provided speaker labels. Zoom is excluded from the comparative table because its platform-integrated participant labels were evaluated using a different attribution procedure.

\begin{table}[!t]
\centering
\small
\begin{tabular}{lc}
\hline
\textbf{System} & \textbf{DER (\%)} \\
\hline
MOSS base    & 16.27 \\
MOSS adapter & \textbf{10.66} \\
Muse         & 19.12 \\
\hline
\end{tabular}
\caption{Pooled system-provided DER for long-form ASR systems, weighted by reference speech time. Values are percentages; lower is better.}
\label{tab:native_der}
\end{table}

The MOSS adapter obtained the lowest system-provided DER among the long-form systems and outperformed MOSS base in every session eligible for both configurations. Muse, despite obtaining the lowest transcription-error point estimates, produced the highest system-provided DER in this comparison. Transcription accuracy and system-provided speaker attribution therefore resulted in different system rankings. Because circular outputs were excluded, however, the pooled DER values are not based on identical sets of recordings and should be interpreted cautiously. Note that MOSS attributes speakers in the same pass as transcription, whereas Whisper produces no speaker labels and depends on an external diarizer.

Zoom yielded a speaker-attribution error of 2.64\% under its platform-specific evaluation. This result is reported as a descriptive platform reference and is not directly comparable with the DER values in Table~\ref{tab:native_der}, because Zoom uses platform-integrated participant information and was evaluated using a different procedure.

Five external acoustic diarizers produced pooled DERs ranging from 12.04\% to 18.91\%, with Nemotron-3-Diarization obtaining the lowest error (Appendix~\ref{app:external_diarization}). These values were all numerically higher than the MOSS adapter's system-provided DER but are not directly comparable because they were calculated using different output boundaries and sets of eligible recordings.

\section{Conclusion}
\label{sec:conclusion}

This study compared six ASR configurations on spontaneous online Spanish L2 interaction. Muse obtained the lowest overall WER point estimate at 15.59\%, closely followed by the MOSS adapter at 16.39\%. The 0.80-percentage-point difference is small relative to the session-level variability of both systems, so we do not read it as establishing an ordering. In-domain LoRA adaptation reduced MOSS WER by 38.2\% relative, bringing the open-weight, long-form system close to commercial Muse, whereas Whisper adaptation produced considerably smaller gains. The effect was confirmed by learner-disjoint cross-validation and came from matched data rather than volume: adding larger out-of-domain bilingual or learner corpora never improved on the classroom sessions alone.

Muse and the MOSS adapter led across speaker roles and language contexts, but substantial difficulty persisted: learner speech produced higher WER than instructor speech for every system, and code-switched and English-only speech were harder than Spanish-only speech. Condition-aware is therefore valuable because it tests whether aggregate advantages generalize and shows where errors remain concentrated.

Transcription and speaker attribution also produced different rankings. Muse provided the lowest transcription-error point estimates, whereas the MOSS adapter obtained the lowest system-provided DER among the evaluated long-form systems, although DER comparisons were based on partially different sets of eligible recordings. For applications that measure participation, the system with the lowest WER is therefore not necessarily the one that produces the most usable transcript.An adaptable open-weight model that runs locally and attributes speakers in the same pass is a practical alternative to a commercial service for classroom data. System selection for bilingual educational data should thus weigh aggregate accuracy, robustness across the intended speakers and language contexts, and the accuracy of speaker attribution.

\section*{Acknowledgments}
We thank Sandra Veigel and Megan Michelle Smith, Who contributed to the quality of transcriptions. This work would not have been possible without their patient and careful effort.

\section*{Limitations}

The evaluation contains seven sessions from one institutional context and one tutoring platform and is limited to online Spanish--English interaction. System comparisons are affected by segmentation differences: Whisper processed reference-defined turns, MOSS and Muse processed longer audio, and Zoom had access to platform-level participant information.

The adapter variants and external-data mixtures were compared on two development sessions, which can rule an approach out but cannot rank configurations that differ by less than half a point, and no full fine-tuning baseline was run. The cross-validation results in Appendix~\ref{app:cv} were scored verbatim on complete recordings and are not directly comparable with the seven-session results.

\section*{Ethical Considerations}

\paragraph{Data and consent.}
The study received institutional IRB approval before data collection. All instructors, tutors, and learners provided informed consent for recording and research use, and only consenting participants were included. Names and other direct identifiers were replaced with anonymized speaker labels.


\paragraph{Privacy and data availability.}
To protect participant privacy, the audio and video recordings are not publicly released because they contain identifiable voices and interactions. We release the transcription guidelines, scoring scripts, and language-classification resources in the supplementary material (Appendix~\ref{app:supplementary}).The classroom transcripts and as-
sociated annotations are restricted; research access may be
considered subject to institutional approval.
They contain no audio or transcripts. AI-assisted writing tools (e.g., ChatGPT and Grammarly) were used solely for grammatical refinement and clarity of presentation.



\bibliography{custom}

\appendix

\clearpage
\section*{Appendix}
\appendix
\raggedbottom

\section{Training and Adaptation Details}
\label{app:adaptation}

This appendix completes the training configuration of Section~\ref{sec:asr_systems} and reports the development results behind it; the recipe was fixed before the seven evaluation sessions were scored. The values below are not comparable with those in Section~\ref{sec:results}: they cover the 38 adaptation sessions, fillers were retained, every system decoded complete recordings, speaker role came from the speaker mapping of concatenated minimum-permutation WER (cpWER; \citealp{vonneumann2023meeteval}) rather than from reference labels and each session was transcribed by the fold-specific adapter that had not seen it.

\subsection{Training Configuration}
\label{app:training_config}
Both adapters used default LoRA initialization without rank-stabilized scaling, optimized with AdamW in bfloat16. A MOSS example was a complete recording, split at silence gaps into parts of at most 30 minutes, with batch size 1, no gradient accumulation, linear decay to zero, no warm-up, gradient checkpointing, float32 adapter weights, and seed 42. A Whisper example was a window of at most 30 seconds of contiguous audio, cut only at gaps between turns and closed when a gap exceeded 5 seconds. Its target used the native timestamp format, with the language token fixed to Spanish in training and detected at inference. Whisper used 24 windows per optimizer step (batch size 12, gradient accumulation 2), cosine decay after a 10\% warm-up, weight decay 0.02, gradient clipping at 0.5, and seed 0. Six epochs were budgeted for both models. For MOSS, we kept the epoch checkpoint with the lowest development WER (ties broken by CER), discarding checkpoints that covered less than 90\% of the audio or whose output length fell outside 0.8 to 1.3 times the reference. For Whisper, the development set was scored at fixed intervals, with early stopping after five evaluations without improvement. The final MOSS adapter was trained on the 57 training examples of the full 38-session set for 264 steps, the budget selected by four of the five cross-validation folds (Appendix~\ref{app:cv}). The final Whisper adapter is the step-275 checkpoint of a 330-step run, 330 being the mean of the best steps across the five folds. With no development session left, it was scored every 55 steps on 200 windows sampled from its own training data, and the lowest window WER was kept: 14.31\% at step 275 against 14.47\% at step 330.

\subsection{Cross-Validated Adaptation Effect}
\label{app:cv}

Five folds were formed over the 38 adaptation sessions so that no learner appeared in both the training and the test portion of a fold; each fold trained on approximately 28 sessions, selected its checkpoint on two development sessions, and transcribed the sessions it had not seen. Each fold trained for six epochs; four folds selected the final checkpoint (252 to 264 steps) and one selected epoch 4 (176 steps), with less than one point of development WER between the two in every fold. Differences are reported with a paired session-level bootstrap of 10,000 replicates \citep{koehn2004significance}. Adaptation reduced MOSS WER from 20.99\% to 17.85\% (Table~\ref{tab:cv_main}), the paired interval excluded zero, and 31 of the 38 sessions improved. The reduction was concentrated in learner speech, 15.92 points against 2.11 for instructor speech, and cpWER fell from 24.91\% to 19.65\% without any diarization-specific training. When a model stopped generating before the end of a recording, the remainder was decoded again, identically for every system; without this recovery the reduction was 2.80 points with an interval of $[-6.47, 1.05]$, widened by two sessions on which one model stopped early.

\begin{table}[!ht]
\centering
\small
\setlength{\tabcolsep}{4pt}
\begin{tabular}{lcccc}
\hline
 & \multicolumn{2}{c}{\textbf{WER}} & \multicolumn{2}{c}{\textbf{CER}} \\
\cline{2-3}\cline{4-5}
\textbf{Subset} & \textbf{Base} & \textbf{Adapter} & \textbf{Base} & \textbf{Adapter} \\
\hline
All 38 sessions            & 20.99 & \textbf{17.85} & 12.48 & \textbf{10.91} \\
Instructor speech          & 16.04 & \textbf{13.93} &  9.99 & \textbf{8.64}  \\
Learner speech             & 50.50 & \textbf{34.58} & 35.04 & \textbf{22.91} \\
Classroom sessions         & 19.48 & \textbf{17.38} & 11.92 & \textbf{10.76} \\
Language-Tutoring          & 25.70 & \textbf{19.31} & 14.27 & \textbf{11.39} \\
Learner--instructor gap    & 34.46 & \textbf{20.65} & 25.05 & \textbf{14.27} \\
\hline
\end{tabular}
\caption{Cross-validated adaptation of MOSS over the 38 adaptation sessions, comparing the base model with the adapter. Values are percentages, except that the learner--instructor gap is in percentage points; lower is better. The overall WER reduction is 3.14 points, 95\% paired bootstrap CI $[-4.19, -2.15]$.}
\label{tab:cv_main}
\end{table}

\subsection{Adapter Variants}
\label{app:variants}

The learning rate was selected on development data from a grid of five values for Whisper and three for MOSS; $5\times10^{-5}$ was best for both, and for MOSS the highest rate, $10^{-4}$, was worse than no adaptation. On the two development sessions of the first fold, plain LoRA (21.85\% WER at the epoch-4 checkpoint) was compared with seven alternatives, none of which improved on it: a loss weighting learner tokens by 3.1 \citep{byrd2019importance} (21.95\%, widening the learner--instructor gap it was meant to close), an unfrozen projector as in SALMONN \citep{tang2024salmonn} (22.05\%), PiSSA initialization \citep{meng2024pissa} (22.24\%, diverging at the sixth epoch), rank-stabilized scaling \citep{kalajdzievski2023rslora} (23.79\%), encoder-only LoRA (23.20\%), decoder-only LoRA (22.17\%), and (IA)$^3$ \citep{liu2022fewshot} (24.36\%). The decoder-only adapter came within 0.32 points of the full adapter with a third of the trainable parameters, and (IA)$^3$, which adapts about 0.01\% of the parameters, left learner WER unchanged (34.90\% against 34.28\%), so neither less nor more trainable capacity appeared to help. 

\subsection{Language Conditioning and Decoding Regime}
\label{app:conditioning}

Explicit language conditioning did not help either model. A language tag (\texttt{<es>}, \texttt{<en>}, or \texttt{<cs>}, from the turn labels of Section~\ref{sec:language_identification}) at the start of every MOSS training turn raised the cross-validated WER over the 38 sessions, scored without re-decoding of stalled output, from 19.25\% for the plain adapter to 22.48\% (22.05\% without adaptation) and removed most of the learner gain; the auxiliary language-identification (LID) loss of \citet{hillah2024tevienes} on Whisper's language token cost 10.57 points against the unadapted model $[3.85, 17.95]$ and 10.00 against the plain adapter. The decoding regime mattered more than any of these (Table~\ref{tab:cv_regime}). Decoded one 30-second window at a time, the regime in which it was trained, the Whisper adapter improved on the unadapted model in every fold, by 4.54 points pooled and 5.04 points as a mean paired per-session difference $[-7.89, -2.57]$; decoded as complete recordings, which chains windows on the timestamps and language the model predicts for itself, the gain disappeared, while the unadapted model lost almost nothing. MOSS transcribes a complete recording in a single pass and retained its gain. This is why the Whisper adapter shows a gain in Section~\ref{sec:results}, where it processes reference-defined speaker turns, and none under complete-recording decoding.

\begin{table}[!ht]
\centering
\small
\setlength{\tabcolsep}{4pt}
\begin{tabular}{lrr}
\hline
\textbf{System} & \textbf{Window} & \textbf{Complete} \\
\hline
Whisper large-v3              & 23.22 & 23.72 \\
Whisper large-v3 adapter           & \textbf{18.68} & 24.29 \\
Whisper adapter, LID loss & 19.77 & 34.29 \\
\hline
\end{tabular}
\caption{The same Whisper models under two decoding regimes, 38 sessions. Window: each 30-second window decoded independently. Complete: sequential long-form decoding. Values are WER percentages.}
\label{tab:cv_regime}
\end{table}

\subsection{External Training Data}
\label{app:external}

\begin{table}[!ht]
\centering
\small
\setlength{\tabcolsep}{3pt}
\begin{tabular}{lcrr}
\hline
 & & \multicolumn{2}{c}{\textbf{WER}} \\
\cline{3-4}
\textbf{Training data} & \textbf{Ext.} & \textbf{Overall} & \textbf{Learner} \\
\hline
Classroom only (control)     & ---  & \textbf{22.59} & \textbf{26.59} \\
+ SPLLOC, all, long          & 34\% & 23.25 & 30.82 \\
+ SPLLOC, UG, long           & 34\% & 23.25 & 28.21 \\
+ Bangor Miami               & 38\% & 23.55 & 31.13 \\
+ SPLLOC, UG, short          & 49\% & 23.74 & 33.05 \\
+ SPLLOC, UG, short          & 34\% & 23.89 & 32.59 \\
+ SPLLOC, all, long          & 50\% & 24.09 & 34.82 \\
No adaptation                & ---  & 24.21 & 34.28 \\
+ Miami and SPLLOC           & 60\% & 24.72 & 33.36 \\
Learner turns only           & ---  & 27.50 & 34.74 \\
\hline
\end{tabular}
\caption{External training data added to the first fold, scored on its two development sessions and sorted by overall WER (\%); lower is better. Ext.: external share of the training data. All: all course levels; UG: undergraduate learners only. Long: SPLLOC sessions concatenated into 15--28-minute examples; short: as recorded. Learner turns only: instructor speech removed. A dash means no external data was used.}
\label{tab:cv_external}
\end{table}

We tested whether out-of-domain speech of a related kind would help: Bangor Miami, Spanish--English conversation between proficient bilingual speakers \citep{deuchar2014miami}, and SPLLOC, interviews with English-speaking learners of Spanish \citep{mitchell2008splloc}. Table~\ref{tab:cv_external} shows the seven mixtures, each added to the first fold and scored on its two development sessions. None improved on the classroom sessions alone, learner WER was worse in every case, and the mixture with the largest external share was worse than no adaptation. Concatenating the five-minute SPLLOC sessions into blocks of 15 to 28 minutes recovered most of the loss seen with short examples, which also taught the model to stop early; removing instructor speech and training on learner turns only made the model stop transcribing within two epochs. Adapters trained on external data alone did worse still: MOSS adapters trained on SPLLOC only (about 55 hours) obtained 23.56\% and 23.80\% WER on four held-out classroom sessions, against 20.08\% for the classroom-only adapter and 21.05\% without adaptation, a paired difference of $+3.48$ $[2.26, 5.55]$; a Whisper adapter trained on Bangor Miami only (17.6 hours, median clip 1.5 seconds) obtained 30.17\% on the 38 sessions against 23.72\% unadapted. The register of the external speech, rather than its quantity, was the limiting factor: neither corpus contains instructor-led classroom interaction.

\section{Supplementary Material}
\label{app:supplementary}

The code repository (\url{https://github.com/samhithaharish/asr-code-only}) contains: (i) the normalization, WER/CER, and DER scoring scripts; (ii) the language-classification code and word lists; (iii) the transcription guidelines for audio-first transcription; and (iv) a readme file. The classroom transcripts and associated annotations are restricted; research access may be considered subject to institutional approval.

\section{Transcription Guidelines and Reference Verification}
\label{app:reference_appendix}

\subsection{Transcription Guidelines}
\label{app:transcription_guidelines}

The independent audio-first dataset was transcribed directly from approximately 45 minutes of audio. Annotators did not consult the Zoom-generated transcripts or any other ASR output during the initial transcription pass. The central instruction was to \emph{write what was heard without correcting, translating, or smoothing the production}. This procedure prevented the audio-first transcripts from inheriting lexical choices, omissions, segmentation decisions, or speaker labels from Zoom.

Annotators used stable anonymized speaker identifiers such as \texttt{Teacher}, \texttt{Student1}, and \texttt{Student2}, together with an instructor--learner role label. Each row represented one coherent speech turn. Rows were split when the speaker changed and merged when a continuous turn by one speaker had been divided artificially. Timestamps were treated as guides rather than as frame-accurate annotations and were changed only when an incorrect boundary affected speaker attribution or transcription.

The following conventions were used:

\begin{itemize}
    \item \textbf{Verbatim learner speech:} Grammatical variation, word order, non-target forms, comprehensible pronunciation-based forms, and non-standard varieties were retained as produced.

    \item \textbf{Code-switching:} Spanish and English words were retained in the positions in which they occurred and were not translated.

    \item \textbf{Fillers, repetitions, and repairs:} Filled pauses, repeated words, false starts, and self-repairs were preserved.

    \item \textbf{Incomplete words:} An incomplete word was marked with a double hyphen at the point of interruption, as in \textit{empeza-{}-} or \textit{I was go-{}- going}.

    \item \textbf{Unclear speech:} Unintelligible material was marked with double parentheses rather than silently guessed.

    \item \textbf{Overlap:} Both speakers were transcribed when possible, and simultaneous speech was marked as overlap.

    \item \textbf{Non-speech events:} Only a controlled set of labels, including \texttt{\{laugh\}}, \texttt{\{cough\}}, \texttt{\{breath\}}, and \texttt{\{pause\}}, was used.

    \item \textbf{Numbers and contractions:} Numbers were written as spoken. Spoken contractions such as \textit{I'm}, \textit{it's}, and \textit{gonna} were preserved.
\end{itemize}

When a turn boundary or speaker identity remained uncertain, the annotator retained the closest defensible segmentation and documented the uncertainty rather than silently normalizing it.

\begin{figure*}[!t]
\centering
\includegraphics[width=\textwidth]{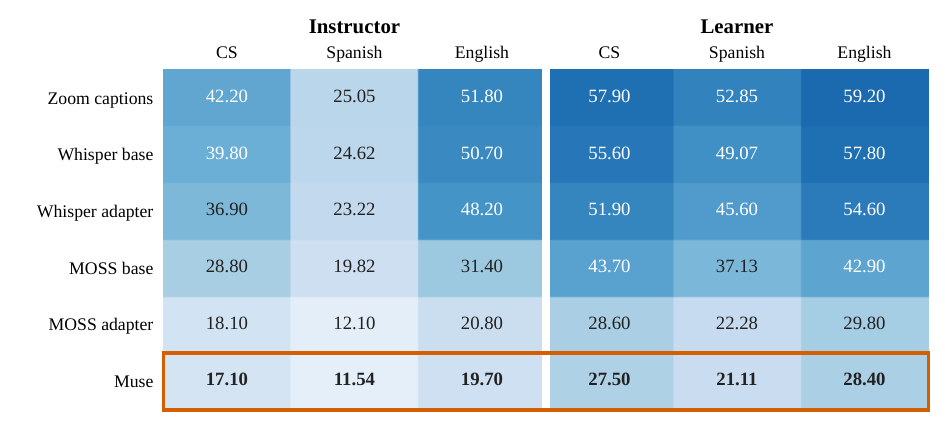}
\caption{WER (\%) jointly disaggregated by speaker role and language context for all six systems. CS denotes code-switched speech. Darker cells indicate higher error, and the lowest value in each column is in bold. The outlined row is Muse Voice Transcribe. Lower is better.}
\label{fig:role_language_heatmap}
\end{figure*}

\subsection{Verification of Human-Corrected Transcripts against Audio-First Transcripts}
\label{app:reference_verification}

The primary evaluation references are human-corrected Zoom transcripts. To verify that constructing them by correcting Zoom output did not introduce systematic differences from what was actually spoken, we compared them against independently produced audio-first transcripts, which were transcribed directly from the audio without consulting Zoom output (Appendix~\ref{app:transcription_guidelines}).

The comparison was restricted to short snippets from the audio-first data, totaling 5.61 minutes of audio and drawn from four different transcripts. For each snippet, the human-corrected transcript was aligned to the audio-first transcript at the word level, and agreement was defined as

\[
\mathrm{Agreement}
=
\left(
1-\frac{\sum (S+D+I)}{\sum N}
\right)\times 100,
\]

where $S$, $D$, and $I$ denote the substitutions, deletions, and insertions of the human-corrected transcript relative to the audio-first transcript, and $N$ is the number of words in the audio-first transcript. Counts were summed over all snippets before the ratio was computed, so that longer snippets contribute proportionally more than shorter ones.

Both transcripts were normalized with the same procedure as the main evaluation (Section~\ref{sec:asr_scoring}) and aligned with word-level edit distance. The snippets, both transcript versions, and per-snippet substitution, deletion, and insertion counts are available on request only.

The resulting pooled agreement was 91.98\%. Because this check covers only 5.61 minutes of audio, it should be read as a verification of reference quality rather than as an exhaustive comparison of the two transcript sets. The main analyses use the human-corrected Zoom transcripts because their coverage of all evaluation sessions supports more stable comparisons by speaker role and language context.

\section{Filler Inventory}
\label{app:supplementary_analyses}
\label{app:filler-inventory}

Fillers were grouped into canonical forms and their orthographic variants. The canonical inventory was
\{\texttt{ahm}, \texttt{ah}, \texttt{uh}, \texttt{eh}, \texttt{erh}, \texttt{emm}, \texttt{hm}, \texttt{mhm}, \texttt{uhuh}, \texttt{oh}\}.
Variants were mapped to their corresponding canonical form (e.g., \texttt{aah} $\rightarrow$ \texttt{ah}) before scoring. The canonical fillers and their variants were then removed from both the reference and hypothesis transcripts.

\section{Joint Disaggregation by Speaker Role and Language Context}
\label{app:role_language_wer}

Figure~\ref{fig:role_language_heatmap} jointly disaggregates WER by speaker role and language context. The instructor conditions contain 132 code-switched turns (2,104 reference words), 1,000 Spanish-only turns (9,640 words), and 76 English-only turns (450 words). The corresponding learner conditions contain 119 code-switched turns (982 words), 635 Spanish-only turns (3,424 words), and 89 English-only turns (266 words).

The joint disaggregation preserved the overall system ranking. Muse obtained the lowest WER in all six conditions, followed closely by the MOSS adapter. Instructor Spanish-only speech produced the lowest error rates, whereas English-only and learner code-switched speech were generally the most challenging conditions. The particularly small English-only cells should be interpreted cautiously.

Adaptation improved all six conditions for both model families. MOSS adaptation reduced WER by 7.72--10.70 percentage points across the instructor conditions and by 13.10--15.10 points across the learner conditions. Whisper adaptation produced smaller reductions of 1.40--3.70 points. The larger absolute reductions in some learner conditions partly reflect their higher baseline error rates and should not be interpreted as evidence of a selective learner benefit.

Because speaker role, language context, and session composition co-vary in this naturalistic corpus, these descriptive results do not constitute a formal statistical test of an interaction between role and language.

\section{External Acoustic Diarization}
\label{app:external_diarization}

Table~\ref{tab:external_der} reports pooled DER for the five external acoustic diarizers.

\begin{table}[!ht]
\centering
\small
\begin{tabular}{lc}
\hline
\textbf{Diarizer} & \textbf{DER (\%)} \\
\hline
pyannote-3.1           & 15.82 \\
pyannote-community-1     & 15.44 \\
DiariZen               & 12.31 \\
NeMo Sortformer        & 18.91 \\
Nemotron-3-Diarization & \textbf{12.04} \\
\hline
\end{tabular}
\caption{Pooled DER for external acoustic diarizers, weighted by reference speech time. Values are percentages; lower is better.}
\label{tab:external_der}
\end{table}

Nemotron-3-Diarization obtained the lowest pooled DER at 12.04\%, followed closely by DiariZen at 12.31\%. Both outperformed the two pyannote configurations, while NeMo Sortformer produced the highest DER at 18.91\%.

\end{document}